\documentclass[letterpaper, 10 pt, journal, twoside]{ieeetran}

\IEEEoverridecommandlockouts

\let\labelindent\relax

	\usepackage{amsmath}
	\usepackage{amsfonts}
	\usepackage{amssymb}
	\usepackage{graphicx}
	\usepackage{cancel}
	\usepackage{mathtools}
	\usepackage{tabularx}
	\usepackage[dvipsnames]{xcolor}
	\usepackage{listings}
	\usepackage{textcomp}
	\usepackage{mathrsfs}
	\usepackage{bbm}
	\usepackage{tikz}
	\usepackage{tikz-cd}
	\usepackage{enumitem}
	\usepackage{arydshln}
	\usepackage{relsize}
	\usepackage{multirow}
	\usepackage{scalerel}
	\usepackage{upgreek}
	\usepackage{ifthen}
	\usepackage{yhmath}
	\usepackage{blkarray}
	\usepackage{dashrule}
	\usepackage{subcaption}
	\usepackage{overpic}
        \usepackage{optidef}

		\newcommand{\abs}[1]{\left|#1\right|}

		\newcommand{\norm}[1]{\abs{\abs{#1}}}

		\newcommand{\paren}[1]{\left(#1\right)}

		\DeclareMathAlphabet{\mathsfit}{T1}{\sfdefault}{\mddefault}{\sldefault}
		
        \newcommand{\track}{no}

        \newcommand{\change}[1]{%
            \ifx\track\no
                \textcolor{black}{#1}%
            \else
                \textcolor{black}{#1}%
            \fi
        }

	\usepackage{letltxmacro}
	\LetLtxMacro\orgvdots\vdots
	\LetLtxMacro\orgddots\ddots

	\makeatletter
	\DeclareRobustCommand\vdots{%
		\mathpalette\@vdots{}%
	}
	\newcommand*{\@vdots}[2]{%
		\sbox0{$#1\cdotp\cdotp\cdotp\m@th$}%
		\sbox2{$#1.\m@th$}%
		\vbox{%
			\dimen@=\wd0 %
			\advance\dimen@ -3\ht2 %
			\kern.5\dimen@
			\dimen@=\wd2 %
			\advance\dimen@ -\ht2 %
			\dimen2=\wd0 %
			\advance\dimen2 -\dimen@
			\vbox to \dimen2{%
				\offinterlineskip
				\copy2 \vfill\copy2 \vfill\copy2 %
			}%
		}%
	}
	\DeclareRobustCommand\ddots{%
		\mathinner{%
			\mathpalette\@ddots{}%
			\mkern\thinmuskip
		}%
	}
	\newcommand*{\@ddots}[2]{%
		\sbox0{$#1\cdotp\cdotp\cdotp\m@th$}%
		\sbox2{$#1.\m@th$}%
		\vbox{%
			\dimen@=\wd0 %
			\advance\dimen@ -3\ht2 %
			\kern.5\dimen@
			\dimen@=\wd2 %
			\advance\dimen@ -\ht2 %
			\dimen2=\wd0 %
			\advance\dimen2 -\dimen@
			\vbox to \dimen2{%
				\offinterlineskip
				\hbox{$#1\mathpunct{.}\m@th$}%
				\vfill
				\hbox{$#1\mathpunct{\kern\wd2}\mathpunct{.}\m@th$}%
				\vfill
				\hbox{$#1\mathpunct{\kern\wd2}\mathpunct{\kern\wd2}\mathpunct{.}\m@th$}%
			}%
		}%
	}
	\makeatother

	\tikzset{
	  symbol/.style={
		draw=none,
		every to/.append style={
		  edge node={node [sloped, allow upside down, auto=false]{$#1$}}}
	  }
	}

\usepackage[bookmarks=true]{hyperref}

\usepackage{cleveref}
\Crefname{figure}{Figure}{Figures}
\Crefname{table}{Table}{Tables}
\Crefname{equation}{Eq.}{Eqs.}
\Crefname{section}{Section}{Sections}
\Crefname{subsection}{Subsection}{Subsections}
\Crefname{appendix}{Appendix}{Appendices}

\usepackage[commentColor=black]{algpseudocodex}
\tikzset{algpxIndentLine/.style={draw=black}}
\algrenewcommand{\alglinenumber}[1]{\bfseries\footnotesize #1}
\algrenewcommand{\textproc}{}
\algrenewcommand{\algorithmicrequire}{\textbf{Input:}}
\algrenewcommand{\algorithmicensure}{\textbf{Output:}}
\usepackage{algorithm}
\makeatletter
\newcommand{\algorithmname}{\ALG@name}
\renewcommand{\floatc@ruled}[2]{{\@fs@cfont #1:} #2\par}
\makeatother

\DeclareMathOperator{\SO}{SO}

\usepackage[letterpaper, lmargin=54pt, tmargin=54pt, rmargin=54pt, bmargin=54pt]{geometry}
\usepackage[suppress]{color-edits}
\addauthor{sg}{magenta}
\addauthor{tc}{cyan}
\addauthor{rt}{teal}
\addauthor{ss}{orange}

\title{Planning Shorter Paths in Graphs of Convex Sets by Undistorting Parametrized Configuration Spaces
}

\author{
Shruti Garg$^{1}$, Thomas Cohn$^{1}$, and Russ Tedrake$^{1}$

\thanks{Manuscript received: November, 11, 2024; Revised February, 26, 2025; Accepted March, 30, 2025.}
\thanks{This paper was recommended for publication by Editor Lucia Pallottino upon evaluation of the Associate Editor and Reviewers' comments.
This work was supported by Amazon.com, PO No. 2D-06310236, Lincoln Labs,  MIT EECS Advanced Undergraduate Research Opportunities Program (SuperUROP) and the National Science Foundation Graduate Research Fellowship Program under Grant No. 2141064.} 
\thanks{$^{1}$Computer Science and Artificial Intelligence Laboratory (CSAIL), Massachusetts Institute of Technology, 32 Vassar St, Cambridge, MA, 02139 {\tt\footnotesize [sgrg,tcohn,russt]@mit.edu}}%
\thanks{Digital Object Identifier (DOI): see top of this page.}
}

\begin{document}

\maketitle

\markboth{IEEE Robotics and Automation Letters. Preprint Version. Accepted April, 2025}
{Garg \MakeLowercase{\textit{et al.}}: Undistorting Parametrized Configuration Spaces}

\begin{abstract}
Optimization based motion planning provides a useful modeling framework through various costs and constraints. Using Graph of Convex Sets (GCS) for trajectory optimization gives guarantees of feasibility and optimality by representing configuration space as the finite union of convex sets. Nonlinear parametrization can be used to extend this technique (to handle cases such as kinematic loops), but this often distorts distances such that convex objectives yield paths suboptimal in the original space. We present a method to extend GCS to nonconvex objectives, allowing us to ``undistort'' the optimization landscape while maintaining feasibility guarantees. We demonstrate our method's efficacy on three different robotic planning domains: a bimanual robot moving an object with both arms, the set of 3D rotations using Euler angles, and a rational parametrization of kinematics that enables certifying regions as collision free. Across the board, our method significantly improves path length and trajectory duration with only a minimal increase in runtime.
\end{abstract}

\section{Introduction}
\label{sec:introduction}
\IEEEPARstart{R}{eliable} motion planning is essential to developing and deploying robotic manipulation systems. Such systems need to produce efficient paths while obeying various constraints. Optimization-based motion planning, which minimizes an objective function while satisfying constraints, offers a powerful paradigm to solve this problem. The decision variables describe the robot's trajectory, the objective allows for choosing desired qualities in the solution, and constraints on these decision variables define obstacle avoidance, dynamic limits, and other interesting task-specific constraints such as coordinating arms in a bimanual system. However, the power of these techniques is tempered by the need to carefully formulate the optimization problems for reliability.

A manipulator’s configuration space is often inherently nonconvex, and nonconvex trajectory optimization usually cannot guarantee optimality (due to local minima), or even feasibility. These guarantees are key to efficient and robust systems that can be used for repetitive motions in safety critical settings. To get a convex optimization formulation that allows for such guarantees, Graphs of Convex Sets (GCS) \cite{marcucci,gcstrajopt} encodes the nonconvexities from obstacle avoidance as discrete decisions. Specifically, the inner approximation of planning or configuration space is represented as a series of intersecting collision free convex subsets. Then, constructing a graph (where vertices are a convex c-space set and edges connect intersecting sets) allows for searching discrete paths through these sets while simultaneously optimizing for the optimal continuous path within each set. Many relevant properties of the trajectory and its derivatives can be transcribed into convex costs and constraints~\cite{gcstrajopt}.

\begin{figure}[t]
    \centering
    \begin{subfigure}[b]{0.73\linewidth}
        \centering
        \includegraphics[width=\linewidth]{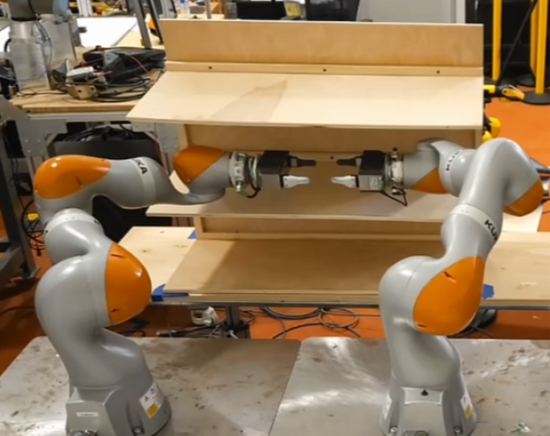}
    \end{subfigure}
    \hfill
    \begin{subfigure}[b]{0.73\linewidth}
        \centering
        \includegraphics[width=\linewidth]{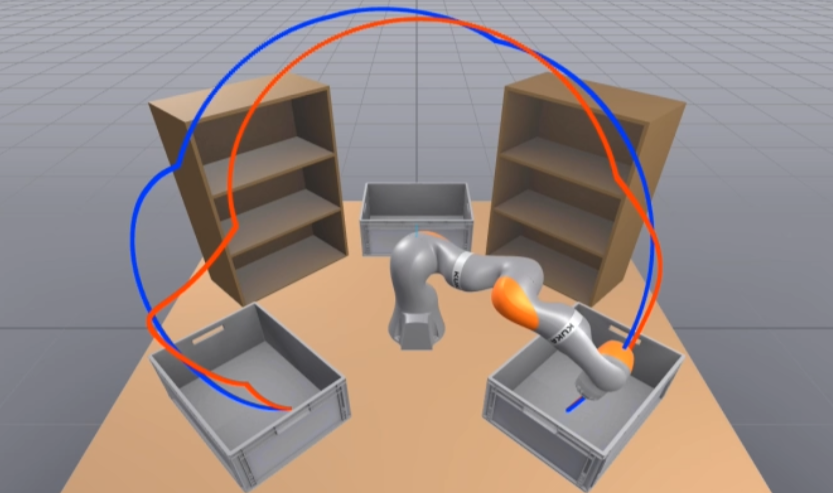}
    \end{subfigure}
    \caption{Experiments include constrained bimanual motion planning between shelves (top) and certifiable 7DoF KUKA iiwa tr\change{aj}ectories between bins (\change{bottom}). The red path is the original result, and the blue path is our improved result.
    }
    \label{fig:setup}
\end{figure}


\change{When the configuration space does not admit finite Euclidean inner approximations, it may be possible under a change of coordinates. For example, end-effector constraints in bimanual robots such as two hands rigidly attached to the same object require planning on a nonlinear manifold in configuration space. Analytic inverse kinematics (IK) enables parameterizing convex sets on this manifold \cite{constrained_bimanual}. Planning over the space of 3D rotations using GCS requires a Euler angles parametrization~\cite[\S2.7.5]{ggcs_thesis}. Even in cases where c-space admits inner approximations, \change{parameterization} can enable useful functionality or properties. For example, using the half tangent rational parametrization to write robot kinematics enables rigorous algebraic collision free certification~\cite{ciris,amice2023certifying}.} 

\change{While} these parametrizations \change{enable GCS to solve key robotics problems, they are also non-isometric}; the shortest path in the parametrized space may not be the shortest path in the configuration space. This distortion leads to suboptimal results when the convex objective used in the parametrized space is a weak approximation for the true objective. \change{Handling nonconvex objectives in GCS}, such as the true objective from the original space, gives more modeling freedom \change{and widens} the breadth of problems we can tackle.

\change{This work's key contribution is using Projected Gradient Descent to optimize nonconvex objectives in a GCS setting.} A gradient based solver guarantees local optimality around the initial guess if the objective is Lipschitz-continuous. We keep constraints convex, so a small convex program can project infeasible solutions back to feasibility. By exploiting structured nonconvexity, \emph{our solver improves optimality of solutions while maintaining feasibility guarantees}. 

For each of the three parametrizations mentioned earlier (bimanual IK, Euler angles, rational kinematics), we formulate and test a nonconvex objective against a convex surrogate in the GCS formulation. This nonconvex optimization is treated as a post-processing step, improving the best solution from GCS. Our method offers significant quantitative and qualitative improvements to motion plans across multiple experiments: path lengths and trajectory times shorten and visual artifacts of planning in the distorted parametrized space are undone. Expanding beyond just countering distortions, we also optimize over a general nonconvex cost, spatial curvature, to speed up bimanual trajectories.

In the rest of \change{the} paper, we review related work on nonconvex trajectory optimization and give necessary background on GCS and nonconvexity \change{in GCS}. Then, we describe our methodology, relevant implementation details, and experimental setups. Finally, we \change{show} results, and conclude with a brief discussion of limitations of our work and potential directions for future research.

\section{Background and Related Work}
\label{sec:related_work}
Sampling-based planners such as Probabilistic Roadmaps \cite{PRM} and Rapidly Exploring Random-Trees \cite{RRT} work very well in practice for kinematic planning problems. By growing finer approximations of the configuration space through sampling, they will eventually find a solution if one exists. Some methods, such as RRT$^{*}$\cite{rrtstar}, can even achieve asymptotic optimality. However, these planners on their own struggle to handle more complex objectives.

Using optimization to solve for the entire trajectory enables more modeling freedom. Objectives can be used to prioritize choice qualities (such as distance or speed) and general constraints are essential for handling dynamics. Roboticists implement optimization based motion planning through a variety of different formulations, including direct collocation~\cite{directcollocation}, Augmented Lagrangian~\cite{augmented_langrangian}, and pseudo-spectral methods~\cite{pseudo_spectral}. These transcriptions can then be solved using general purpose solvers such as SNOPT~\cite{snopt} or gradient-based methods. For example, KOMO uses gradient descent on an Augmented Lagrangian transcription~\cite{komo}, and CHOMP uses covariant gradient descent~\cite{chomp}. Nonconvexity will often be handled with clever initialization or stochasticity, such as in STOMP~\cite{stomp}. cuRobo~\cite{curobo_icra23} moves all constraints into the objective and leverages parallelization to simultaneously consider many initial guesses. Across all of these approaches, the formulation remains nonconvex, lacking feasibility or optimality guarantees.

\subsection{Graphs of Convex Sets}
\label{sec:methodology:gcs}
Graph of Convex Sets (GCS) presents a new strategy for solving \change{the shortest path problem} with continuous and discrete decisions. Formally, a GCS is a graph, where each vertex $v$ has an associated continuous variable $x_v$ within a convex set $X_v$, and each edge $(u,v)$ is a convex function of $x_u$ and $x_v$. Finding the shortest path $\mathcal P$ through this graph can then be formulated as a Mixed Integer Convex Problem (MICP) with $\mathcal P$ and $x_v$ as decision variables~\cite[\S5]{marcucci}.



\change{GCS can be used to solve motion planning problems when given convex collision-free sets. These sets constitute the vertices of $\mathcal P$, and their union is an inner approximation of our planning space. The decision variable $x_v$ describes the continuous trajectory through the set $v$. As in GcsTrajOpt~\cite{gcstrajopt}, we define $x_v$ as the control points of a B\'ezier curve to parametrize the continuous trajectory.} This choice admits convex path continuity and differentiability constraints, and \change{guarantees} collision-avoidance of the whole trajectory~\cite[p.9]{gcstrajopt}. GcsTrajOpt minimizes the distance between adjacent control points as a proxy for minimizing the length of the curve~\cite{marcucci}. We use the same objective for the convex relaxation and any convex optimization we do.

The above discussion focuses only on the path, but in GcsTrajOpt, $x_v$ also has a time-scaling variable, $h$. \change{We want to avoid nonconvex acceleration constraints that use this time parametrization variable.} Instead, we use 
\change{TOPP-RA (Time Optimal Path Parametrization based on Reachability Analysis)~\cite{toppra}} to generate timed trajectories from our planned spatial paths. \change{Any differentiable path can be navigated under acceleration constraints by slowing down, so using TOPP-RA maintains feasibility guarantees by keeping nonconvexity out of our constraints.}  So, for a collision free convex set $Q_i$, our vertices are of dimension $Q_i^{d+1}$ given B\'{e}zier curves of degree $d$ with $d+1$ control points.

\subsection{Parametrizing Configuration Space}
\label{sec:backg:parametrization}

In some cases the configuration space benefits from being parametrized to enable building convex sets for GCS or generating collision free certificates. In this sub-section we review three such parametrizations and associated related works that apply GCS to manipulation motion planning problems. These parametrizations form the three main cases we will tackle with our method in the rest of the paper.

\subsubsection{IK and Constrained Bimanual Planning}
\label{sec:backg:ik}
Constrained bimanual manipulation, \change{or when} two robot arms move with a fixed transform between their end effectors, requires a equality constraint in task space. \change{This non-linear inequality prevents us from using GCS on the $\mathbb{R}^{14}$ configuration manifold as it is. }Cohn et al.~\cite{constrained_bimanual} use \change{IK} to determine the joint angles of a subordinate robot given the end effector position of the leading arm. This parametrization collapses the $\mathbb{R}^{14}$ full joint space into an $\mathbb{R}^{8}$ space formed by the leading arm's joints and a redundancy factor (required to generate consistent joint angles for the subordinate arm). However, \change{minimizing path length in the $\mathbb{R}^{8}$ only minimizes the path length for the leading arm and ignores the subordinate arm.} As a result, paths visibly favor the leading arm.

\subsubsection{Euler Angles and GCS on \texorpdfstring{$\SO(3)$}{SO(3)}}
\label{sec:backg:so3}
Planning over $\SO(3)$, the set of 3D rotations, is an important motion planning domain in robotics. As any roboticist knows, there are many different ways to represent rotations. Rotation matrices perfectly represent $\SO(3)$, but require bilinear constraints when included in optimization problems (constraints to ensure the validity of the Rotation matrix). Cohn et al.~\cite{ggcs_thesis} explores different parametrizations to plan over rotations with GCS: Euler angles, axis-angle, and quaternions. The axis-angle and quaternion representations require piecewise-linear approximations, and also require solving two planning problems due to double cover of $\SO(3)$. Though Euler angles are quicker to plan over, the original work observes planning with Euler angles gives longer paths than the quaternion and axis-angle approximations. This discrepancy is due to the distortion of the underlying geometry of $\SO(3)$: distances get arbitrarily large when approaching gimbal lock. 

\subsubsection{Rational Kinematics to 
Certify Collision Free}
\label{sec:backg:rkin}
The forward kinematic mapping (needed for checking if a configuration is collision-free) is a trigonometric polynomial. Amice et al.~\cite{ciris} \change{write this nonconvex relationship as a} multi-linear polynomial, using the tangent half-angle substitution $s=\tan\frac{\theta}{2}$, further implying
\[
\sin(\theta)=(1-s^2)/(1+s^2),\; \cos(\theta)=(2s^2)/(1+s^2),
\]
for $\theta\in(-\pi,\pi)$. \change{Changing coordinates allows the formulation of Semi-Definite Programs (SDPs) to certify non-collision of regions in the robot's rational c-space with task space obstacles.} This certification can be done for individual convex sets~\cite{ciris}, or even an entire trajectory~\cite{amice2023certifying}. 

\begin{figure}[t]
    \centering
    \begin{subfigure}[b]{0.23\textwidth}
        \centering
        \includegraphics[width=\textwidth]{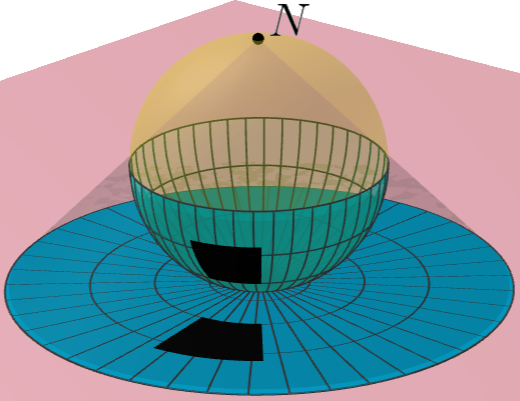}
    \end{subfigure}
    \caption{A stereographic projection about $N$ projects the bottom of the black rectangle as being smaller than the top. An optimal distance planner operating in the post-projection (parametrized) space would favor the bottom despite the sides being equal in actuality. Image generated using \cite{animation-stereo}.}
    \label{fig:stereo}
\end{figure}

The \emph{rational} parametrization of kinematics, similar to the Stereographic Projection, is non-isometric. \change{Most obviously} when $\theta$ approaches $\pm \pi$, $\tan\frac{\theta}{2}$ asymptotically approaches to $\pm \infty$. This means that in the parametrized space, as joint values approach limits at $\pm \pi$, distances grow arbitrarily. More generally, equidistant points in the original space will be closer together in  parametrized space \change{near zero} than the same points further away from zero. Therefore, a convex formulation of distance in the parametrized \change{space} will be inaccurate. Specifically, we expect if any joint is moving near $\pm \pi$ away from the point of stereographic projection, \change{the} paths planned will be sub-optimal due to \change{distortion}.

For all of these cases, convex objectives being minimized in GCS are in the parametrized spaces and therefore subject to the discussed pathologies. The planner clearly would benefit from the use of nonconvex objectives that represent \change{the true objectives in the original space}. This work bridges the gap between using nonconvex objectives and maintaining guarantees of GCS due to convexity.

\subsection{Nonconvexity and GCS}
There is precedent for \change{handling} nonconvexity in GCS or similar optimization frameworks. The original GcsTrajOpt preprint~\cite{gcstrajopt_preprint} suggested using convex approximations to incorporate nonconvex objectives and constraints. In line with this suggestion, existing GCS works using the parametrizations from \Cref{sec:backg:parametrization} use a convex surrogate objective to approximate the optimal solution. While this approach preserves convexity, the approximations are inherently heuristic and must be hand-designed. Moreover, optimizing for an approximation \change{bounds} the optimality of the solution by the quality of the approximation. The nuances of the approximations can also lead to systematic pathologies, such as the imbalance between arms in the bimanual planning domain.

Another approach is to \change{use} local convex approximations of the nonconvexity. Clark and Xie \cite{clark2023planning} suggest approximating the nonconvex costs using piecewise-linear approximations and creating smaller sets within which the objective is convex. This approach maintains convexity, but may scale poorly when dealing with complex objectives and finer approximations. Using a mix of biconvex alternation and local convex approximation, Fast Path Planning \cite{fpp} handles a bilinear in a similar set-up as GCS. The nonconvexity in this problem is contained to the constraints and is handled using alternation. The nonconvexity being a bilinear nonconvexity is key to enabling this method. Our work aims to enable a broader class of nonconvexity in the objective functions.

Enabling GCS to handle nonconvexity without approximation expands the method's applicability and improves solution quality. We restrict ourselves to nonconvexity in only the objective to improve motion planning results while maintaining guarantees. This restriction does prevent us from handling acceleration constraints, due to their nonconvexity in the GcsTrajOpt formulation. Von Wrangel~\cite{davidthesis} presents specific strategies for handling certain common nonconvexities in GCS, including acceleration constraints. But the empirical success comes without strong guarantees.

\section{Methodology}
\label{sec:methodology}
\subsection{Nonlinear Changes of Coordinates}
\label{sec:methodology:nonlinearCoC}

Each of the parametrizations \change{from} \Cref{sec:backg:parametrization} distort the robot's configuration space by introducing a nonlinear change of coordinates. More formally, each domain has a smooth (nonlinear) transformation $\alpha: Q \rightarrow C$ that maps $x$ from the more useful \emph{parametrized space} $Q$ to a point $\alpha(x)$ in the original configuration space $C$. Each of the works used GCS to solve for a trajectory in $Q$, which then is remapped to $C$ using $\alpha$ to get an actual robot trajectory. However, since $\alpha$ is a nonlinear transformation, the minimum length \change{path} in $Q$ is not guaranteed to be the minimum length \change{path} in $C$. 

The key limitation is that the convex path length cost in $Q$ can be arbitrarily far from the true objective: minimizing distance in $C$. Using $\alpha$ \change{in the objective} enables changing coordinates back to the original space $C$ and defining a true (now nonconvex) objective, but the sets and constraints stay convex in the parametrized space $Q$. 

For the constrained bimanual case, \change{we define} $\alpha: \mathbb{R}^{8} \rightarrow \mathbb{R}^{14}$ is as the nonlinear analytic IK function with an original configuration space of both arms' joints ($\mathbb{R}^{14}$) and a parametrized space of one arm's joints and the self-motion of the other arm ($\mathbb{R}^{8}$). For planning over $\SO(3)$ with Euler angles, $\alpha: \mathbb{R}^{3} \rightarrow \mathbb{R}^{4}$ is the standard conversion from Euler angles to quaternions. For planning in the rational parametrization of kinematics, $\alpha$ is defined as $\theta=2\tan^{-1} s$.

\subsection{Formulating the Optimization}
\label{sec:methodology:solver}

The nonconvex objective using $\alpha$ still needs to be expressed in terms of our decision variables $x_v$, the control points of the B\'ezier curve in $Q$. We cannot directly apply $\alpha$ on $x_v$ to define a distance objective as the remapped control points from $Q$ do not define a same B\'ezier curve in $C$. However, any points along the B\'ezier curve in $Q$ will still be along the same path in $C$, and any point along the B\'ezier curve is a convex combination of its control points. Therefore, a piecewise-linear approximation of the curve in $Q$ \change{maps} to a piecewise-linear approximation in $C$ using $\alpha$. 

The representative cost can then be the length of this piecewise-linear approximation. For the bimanual and rational configuration \change{experiments, we} sum the Euclidean distance between each adjacent pair of points in the full configuration space. For $\SO(3)$, we use the length of the Spherical Linear Interpolation (SLERP) path since the underlying geometry is a sphere. For better results, we square the length of each \change{piece. This objective is better numerically for the optimizer and in the limit of an infinitely-fine discretization, it will both produce the same answer as a piecewise L2 norm \cite[p.189]{riemanniangeo}. For our experiments, using 10 samples per region to estimate the path strikes a good balance of accuracy and speed: a higher resolution approximation will be more accurate, but require more computational effort.}

The GcsTrajOpt \cite{gcstrajopt} transcription with the original convex objective in the changed coordinates can be written \change{as the following} where $x_{ij}$ is the $j^{th}$ control point of the B\'ezier curve in set $Q_i$:
\begin{mini*}|s|
{\mathbf{x_v}}{\sum_{i = 0}^{v} \sum_{j = 0}^{d} ||x_{ij}-x_{ij-1}||}
{}{}
\addConstraint{x_i \in Q_i , Q_i \in \mathcal P}
\end{mini*}
Our proposed optimization is:
\begin{mini*}|s|
{\mathbf{x_v}}{\sum_{i = 0}^{v} \sum_{k = 0}^{10} f(\alpha(x_{ik}), \alpha(x_{ik-1}))^2}
\addConstraint{}
\addConstraint{x_i \in Q_i, Q_i \in \mathcal P}
\end{mini*}
where $x_{ik}$ is the $k^{th}$ sampled point in set $Q_i$ and $f(a, b)$ gives the distance between two points $a$ and $b$ in $C$. Note that both optimizations live in $Q$ with collision free sets $Q_i \subseteq Q$ but have different objectives. Thus, our optimization problem is specifically structured to isolate the nonconvexity in the objective function via the parametrization $\alpha$.


\subsection{Projected Gradient Descent}
To exploit the aforementioned structure, we use \emph{Projected Gradient Descent} (PGD) to maintain guarantees of feasibility and optimality. PGD is an iterative first-order or gradient-based solver with two parts: the gradient step and the projection back into feasibility. PGD steps in the direction of steepest decrease of the objective until a minimum is achieved. If any step yields an infeasible configuration, the solver projects the updated point back into feasible space. Because the constraints remain convex, the projection, a quadratic program finding the closest point in the set, solved with Mosek~\cite{mosek}, always returns a solution. Moreover, the convergence of PGD is well-understood if the multiplication factor of the negative gradient is less than or equal to the Lipschitz constant of our objective function~\cite{lipschitz}. Our objective landscapes do not admit Lipschitz constants, but those that do can leverage this useful guarantee.

\subsection{Solver Performance}

Beyond theoretical guarantees, certain implementation details further improve the performance of our solver. 

\subsubsection{Initialization} As PGD finds local minimizers, the solution highly depends on the initialization. Within the GCS workflow, this initialization can come from two distinct candidates: after or during the rounding procedure (the step which projects the convex relaxation result to the near optimal discrete solution). \change{Post-processing} the solution after rounding is a great way to quickly improve    a fixed discrete path in the parametrized space. We focus on this method, but one could also use this nonconvex optimization for each sampled path as an integral component of the rounding stage.

\subsubsection{Optimal Step Sizes} 
Our objectives are too complex to easily identify the Lipschitz constant and theoretically \change{find} a good step size. \change{Classical PGD would then require manually tuning step size, so} we use the backtracking line search PGD~\cite{backtracking}, which searches for an optimal step size. It repeatedly halves an upper bound on the step size till the Armijo condition of sufficient decrease is met. This keeps the solver from overshooting minima while \change{converging fast}. 

\subsubsection{Gradient Precompilation} Initially, gradient computations were \change{most} of the runtime. Precompiling gradients with JAX~\cite{jax} moves this time cost offline \change{to speed} up the PGD iterations. Compiling gradients for each vertex individually also allows us to  re-use computations for \change{start goal pair.}

\subsubsection{Affine Projections} With the gradients pre-compiled and checking for feasibility being fast because our feasible space can be expressed as a halfspace intersection, the majority of the time cost comes from the QP projection step. To reduce the number of QP solves and optimize for speed, we initially project onto the affine hull of the feasibility polyhedron. This projection satisfies any equality constraints such as path continuity and differentiability. This projection is much cheaper than the QP projection, since we can efficiently compute the affine hull. (All equality constraints are known explicitly, and the convex sets making up the GCS are positive volume, since they are produced by the IRIS-NP algorithm~\cite{irisnp}.) In some cases (especially with smaller step sizes), this projection will suffice to push the solution back into feasibility, saving time for the solver. If the point is still infeasible, the solver runs the full QP.

\subsubsection{Convergence Criteria} The solver tracks the moving average of the cost over the last 5 iterations, and terminates when the average changes by less than 0.5\%. The moving average prevents us from terminating early; the cost occasionally jumps for a single iteration before continuing on a significant downward trend. For cases that do not converge, the solver terminates after a maximum of 70 iterations. We hypothesize this occurs when the projection step increases the cost too much, indicating a high Lipschitz constant. In practice for these experiments, optimizations that converged, typically converged well before 70 iterations.

\subsection{More general nonconvex objectives: Curvature}
\label{sec:methodology:curvature}

So far the methodology has focused primarily on the special case of eliminating the distortion caused by non-isometric parametrizations. However, we can also optimize for any smooth nonconvex objective, expanding our modeling power. Some examples of useful nonconvex objectives would be minimizing curvature (or other higher-order path derivatives) or penalizing proximity to obstacles.

Penalizing the \emph{curvature} of the path
\[
\kappa=\paren{\norm{x'}^{-3}}\sqrt{\norm{x'}^2\norm{x''}^2-\left(x'\cdot x''\vphantom{\norm{x'}}\right)^2}
\]
should help \change{TOPP-RA} produce better trajectories, as high curvature paths contain tight turns, that require a slower traversal to stay within acceleration limits. Although such paths might be longer than those produced by a pure shortest-path trajectory, they can be traversed more quickly.

We define this objective \change{too} over sampled points along the path defined by $x_v$. Given sampled points, we calculate the curvature of each point and then apply the RealSoftMax (a smooth maximum function) to approximate the maximum curvature of our paths.  We expect paths under this optimization will have higher path length but lower \change{duration} when time-parametrized by \change{TOPP-RA}.

\section{Experiments}
\label{sec:experiments}
In this section, we detail the results collected on the three motion planning domains of interest: constrained bimanual, $\SO(3)$ with Euler angles, and rational kinematics. For all of our experiments, we solve the GCS problem with the original convex objective first and then run the projected gradient descent to improve the solution.
Interactive recordings of all trajectories and other results are available online at \href{https://shrutigarg914.github.io/pgd-gcs-results/}{https://shrutigarg914.github.io/pgd-gcs-results/}
\begin{figure}[b]
    \centering
    \smallskip
    \includegraphics[width=0.85\linewidth]{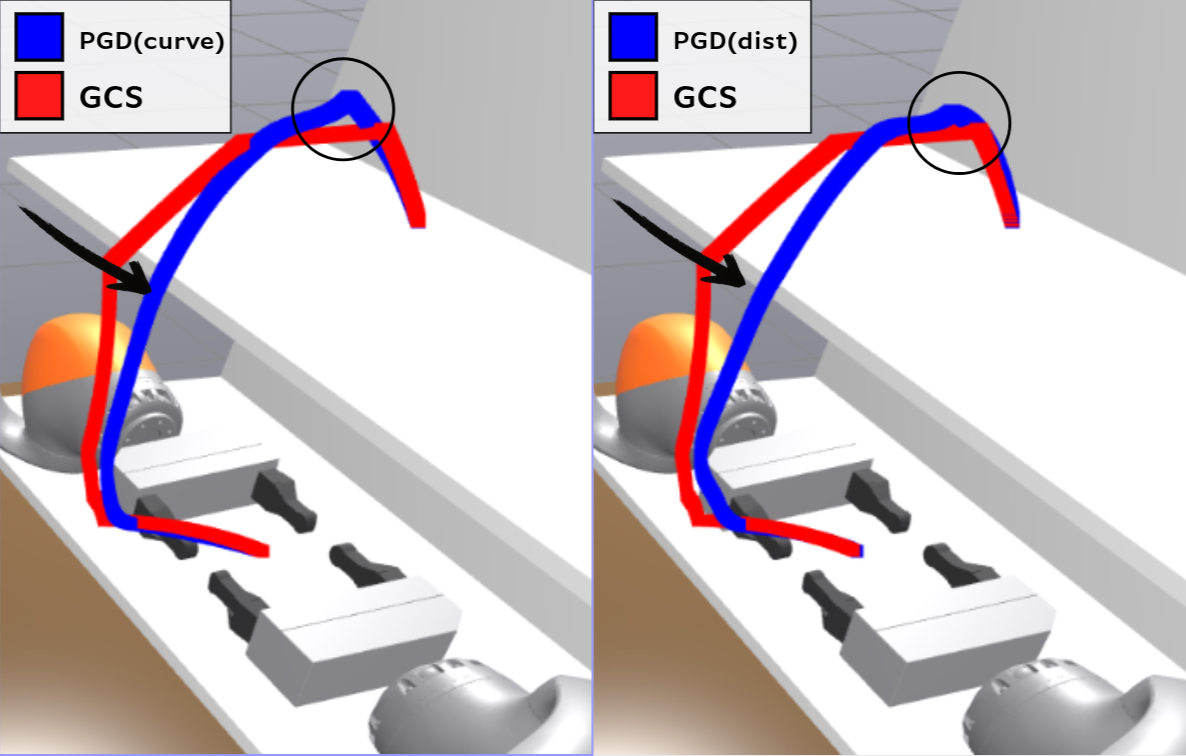}
    \caption{Optimizing jointly for curvature and distance yields quicker trajectories but longer distances--the curvature-regularized path is farther from the shelf.}
    \label{fig:curvature}
\end{figure}

\subsection{Constrained Bimanual Motion Planning}
In this experiment, two iiwas navigate a shelf while keeping the transform between end effectors constant, as if they were jointly carrying an object as shown in \Cref{fig:setup}. We evaluate the PGD solver on the key start and goal pairs from~\cite{constrained_bimanual} both on hardware and in simulation. The comparison of these benchmark paths before and after optimizing for the nonconvex objective is presented in \Cref{tab:topmiddlebottom}. The ``GCS'' column indicates using the convex $\mathbb{R}^{8}$ objective, the ``Distance'' column indicates using the nonconvex $\mathbb{R}^{14}$ objective, and the ``Curvature + Distance'' column indicates \change{using a linear} combination of the nonconvex $\mathbb{R}^{14}$ objective and the nonconvex curvature cost with a ratio of 8 to 0.01 respectively. \change{This ratio is a hand tuned parameter to compensate for the path distance objective being on the order of 100 times greater than the path curvature objective.} While the $\mathbb{R}^{14}$ objective results in the shortest paths, regularizing for lower curvature lengthens paths, but shortens traversal times after \change{TOPP-RA}'s re-timing. Visually, minimizing this joint objective leads to more rounded paths, as shown in \Cref{fig:curvature}.

\begin{figure}[t]
    \centering
    \includegraphics[width=0.88\linewidth]{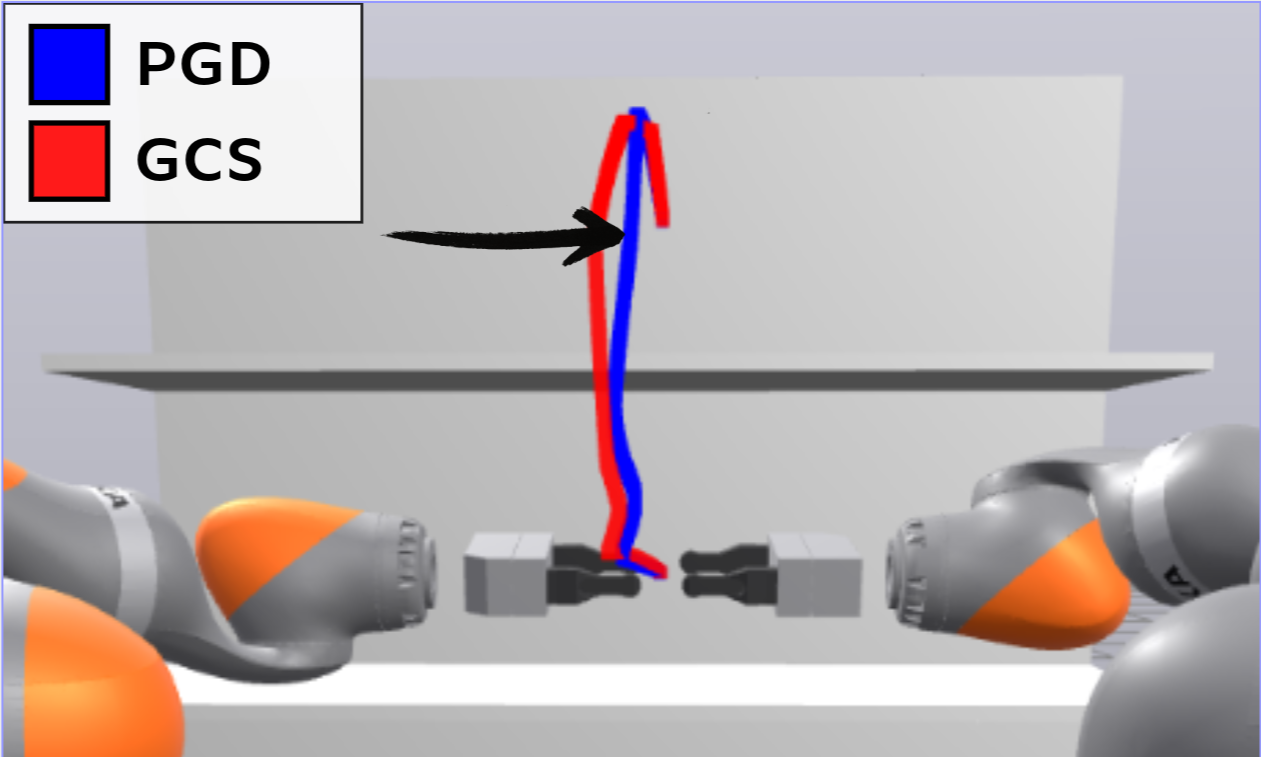}
    \caption{Paths become more centered \change{as the nonconvex objective accounts for the distance traveled by both arms. The original convex objective just accounts for the controlled arm.}}
    \label{fig:pathlengthcomp}
\end{figure}

To quantify the difference in distance traveled between the arms, we define the \emph{imbalance} of a trajectory as $(d_s - d_c)/(d_s + d_c)$, where $d_c$ is the distance traveled by the controlled arm and $d_s$ is the distance traveled by the subordinate arm. When both arms travel comparable distances, the imbalance distribution centers around 0. When one arm travels much longer distributions than the other, the imbalance metric approaches $\pm1$ in magnitude. Table \ref{tab:topmiddlebottom} shows that this imbalance metric approaches 0 after post-processing under the $\mathbb{R}^{14}$ objective. In \Cref{fig:pathlengthcomp}, we see paths favour the leading arm less. \change{The imbalance for jointly optimizing curvature and distance is higher than optimizing just the distance indicating that smoother paths are more imbalanced.} This asymmetry likely comes from the same-handedness of the iiwas.

For a more comprehensive analysis, we randomly sample 100 start and end points from the valid and reachable configuration space. Paths generated are on average 20.60\% shorter in the $\mathbb{R}^{14}$ configuration space after applying our post-processing step. \change{These paths take on average 31.02\% less time to navigate.} The imbalance shifts towards 0, indicating that the paths for the subordinate arm are more comparable to the leading arm after the nonconvex optimization. These improvements took an average of 0.0554 seconds of compute (approximately 13.7 iterations) in addition to the 2.133 seconds that the surrogate convex optimization takes. 

\begin{table}[t]
    \centering
    \smallskip
    \caption{Optimizing over the nonconvex cost improves metrics for the three \change{benchmark} trajectories.
    }
    \label{tab:topmiddlebottom}
    \begin{tabular}{|l|c|c|c|}
        \hline
        \multicolumn{4}{|c|}{Top to Middle}\\
        \hline
        & GCS & Distance & Distance + Curvature\\
        \hline
        Trajectory Time & 4.889 & 3.469 & \textbf{3.243} \\
        \hline
        R14 Path Length & 4.241 & \textbf{3.766} & 3.884 \\
        \hline
        Imbalance & \change{0.331} & \change{\textbf{0.117}} & \change{0.216} \\
        \hline
        \hline
        \multicolumn{4}{|c|}{Middle to Bottom}\\
        \hline
        & GCS & Distance & Distance + Curvature\\
        \hline
        Trajectory Time & 5.326 & 3.08 & \textbf{2.99} \\
        \hline
        R14 Path Length & 3.325 & \textbf{3.175} & 3.247 \\
        \hline
        Imbalance & \change{0.162} & \change{\textbf{0.099}} & \change{0.110} \\
        \hline
        \hline
        \multicolumn{4}{|c|}{Top to Bottom}\\
        \hline
        & GCS & Distance & Distance + Curvature \\
        \hline
        Trajectory Time & 7.48 & 4.263 & \textbf{3.99} \\
        \hline
        R14 Path Length & 5.622 &  \textbf{5.048} & 5.13 \\
        \hline
        Imbalance & \change{0.190} & \change{\textbf{0.084}} & \change{0.122} \\
        \hline
    \end{tabular}
\end{table}

80\% of the runtime is solving QP projections. The affine projection is only useful when step size is fixed. When using backtracking to determine step size, the projection onto the affine hull is almost never sufficient. This observation indicates to us that at our boundary the gradients are consistently pointing outward. This is not unexpected, since the collision-avoidance constraints are active at the boundary, and moving closer to obstacles generally allows a shorter path length.

\subsection{Planning over \texorpdfstring{$\SO(3)$}{SO(3)}}
For our last experiment, we plan over random start and goal 3D rotations independent of a simulation or hardware set-up. To cover SO(3), we set up the same charts and convex regions as in~\cite{ggcs_thesis} for the Euler angles, quaternions, and axis-angle parametrizations. The latter two use piecewise-linear approximations of the original $\SO(3)$ space and act as baselines. We run PGD on the Euler angles setting only. The Euler angles GCS graph is fully connected and optimizes Euclidean distance within each set, so the shortest path between any two points will be a linear path, regardless of the order of our B\'ezier curves. For time-efficiency, we generate order one GCS solutions and initialize the PGD solver with the control points evenly spaced along each line segment.

\begin{figure}[tb]
    \centering
    \begin{subfigure}[b]{\linewidth}
        \centering
        \smallskip
        \includegraphics[width=\linewidth]{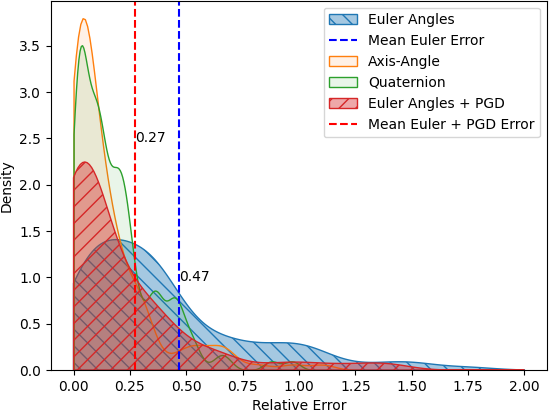}
        \label{fig:trajt_comp}
    \end{subfigure}
   \caption{Comparing the distributions of relative error of paths with respect to the SLERP distance between start and goal orientations. The PGD significantly improves the results of the Euler angles parametrization.}\label{fig:so3_experiment}
\end{figure}

Given there are no obstacles, SLERP gives the shortest path between any two orientations. We use it as the closed-form ground truth distance. \Cref{fig:so3_experiment} shows the distribution of relative error in path length for the three representations of $\SO(3)$ when planning across 125 random start and goal pairs, along with the PGD post-processing on the Euler angles paths. The distribution of error for Euler angles shifts significantly closer to 0 after running PGD. The relative error decreases by 42.5\% on average. This improvement has a bimodal distribution: for some paths the PGD greatly improves the solution, but for others, there is little improvement to be made. The latter might be local minima, where the global optimal lies on a different discrete path.

These improvements on average \change{take} 4.08 seconds \change{in addition to }the 17.28 seconds \change{taken} to generate the original solutions for Euler angles. Of this time, the solver only ran for 0.62 seconds. The remaining 3.46 seconds were \change{spent re-using} vertices and B\'ezier curves from the original solution and could be further optimized. Comparatively, planning with axis-angles \change{takes} 41.10 seconds. At a lower resolution quaternions take 16.73 seconds, but for higher resolutions, their solve time is on the order of minutes. \change{Our method offers a way to generate more accurate paths using} Euler angles while still being faster than the more accurate axis-angle and quaternion representations. \change{Moreover, of the three, only Euler angles allow for  using IRIS-NP~\cite{irisnp} to grow collision-free regions in the presence of obstacles.}

\subsection{Rational Parametrizations of Robot Kinematics}

We have two experimental settings in simulation that use the rational kinematics parametrization. One is a 3 degree-of-freedom iiwa (four of the joints are locked) that moves within a vertical 2D plane. The other is a 7 degree-of-freedom iiwa mounted on a table, as shown in \Cref{fig:setup}. The nominal position (i.e. point of projection) for both iiwas is when the arms stand straight up with all joint angles at 0. All the regions in the 3DoF case are certified to be completely collision-free using the Certified IRIS algorithm~\cite{ciris}. All the trajectories in the 7DoF setting can be certified using~\cite{amice2023certifying}.

\begin{figure}[t]
    \centering
    \smallskip
    \begin{subfigure}[b]{0.25\textwidth}
        \centering
        \includegraphics[width=\linewidth]{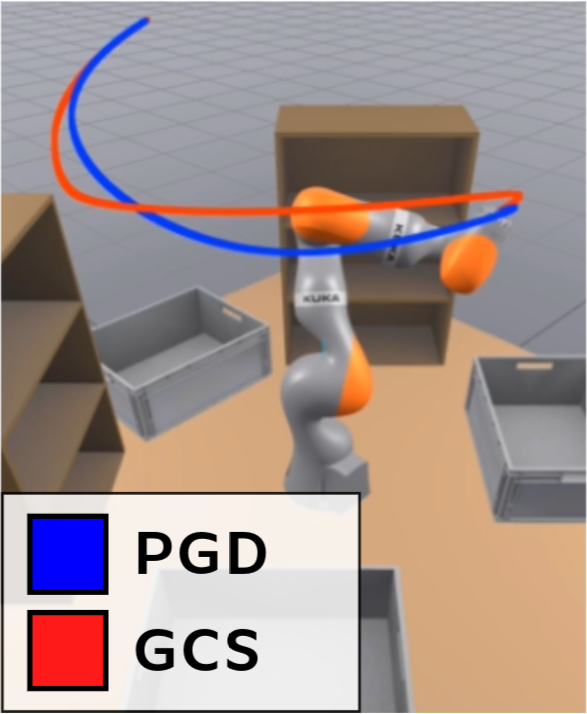}
    \end{subfigure}
    \begin{subfigure}[b]{0.218\textwidth}
        \centering
        \includegraphics[width=\linewidth]{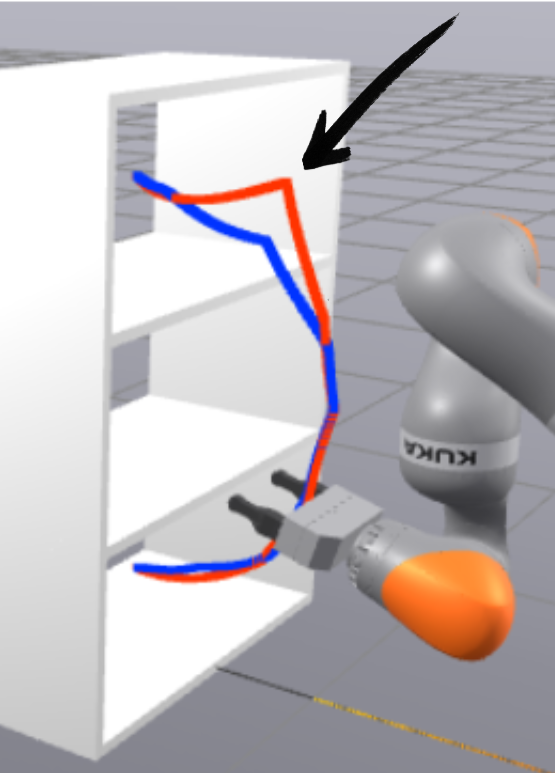}
    \end{subfigure}
    \caption{The 3DoF iiwa (right)  skews towards the nominal position in the original GCS solution (in red). The 7DoF iiwa (left) shows improvement in path  before and after the post-processing for a random start and goal configuration.}
    \label{fig:ciris_results}
\end{figure}

For the 3DoF planar iiwa, \change{qualitatively} the paths become less biased towards the point of projection:  in \Cref{fig:ciris_results}, the the PGD refinement \change{reduces the} extraneous spike towards the nominal pose. Quantitatively, most paths \change{show little} improvement across 100 random start and goal points among the shelves. On average, the paths get 0.2\% shorter and most terminate within 7 iterations and 0.22 seconds. \change{The example in Figure \ref{fig:ciris_results} shows a 1.2\% improvement in path length. Weaker numerical results are expected as the configuration space distorts most intensely near the joint limits, so the average case does not have much room for improvement.}

For the 7dof iiwa, the projected gradient descent on random paths in configuration space between the bins results in 3.89\% shorter in path lengths and a 4.74\% shorter trajectory times. When one or more joints travel near their limits, these improvements are higher. For example, \Cref{fig:ciris_results} shows a trajectory that gets 10.8\% shorter and 17.6\% faster.

\section{Discussion}
\label{sec:discussion}
We have presented a method to solve GCS problems with nonconvex objectives, granting greater modeling freedom and yielding better motion plans. By keeping the constraints convex, we maintain the feasibility guarantees of GCS and avoid the inconsistency typical of nonconvex optimizations.

Our method is particularly effective when accounting for the distortion from nonlinear parametrizations of planning spaces.
In constrained bimanual motion planning, our post-processing step produces paths that are more balanced between the arms, 20\% shorter on average, and 31.02\% faster after being time-parametrized. For Euler angles, the paths are 40\% shorter on average. \change{Beyond undistorting paths, the approach enables optimizing general nonconvex objectives such as curvature. For the bimanual setting, we find paths with greater curvature radii and quicker traversal. The lack of significant change in path length in the average for the rational kinematic case suggests that the distortion from the stereographic projection is not usually significant. Thus, planning in this parametrization of configuration space and enabling rigorous certification plausibly outweighs the minor cost increase. Even then, our method produces strong improvements in the worst case, and in the average case with little room for change, the solver terminates quickly.}
\begin{figure}[b]
    \centering
    \begin{subfigure}[b]{0.48\linewidth}
        \centering
        \includegraphics[width=\linewidth]{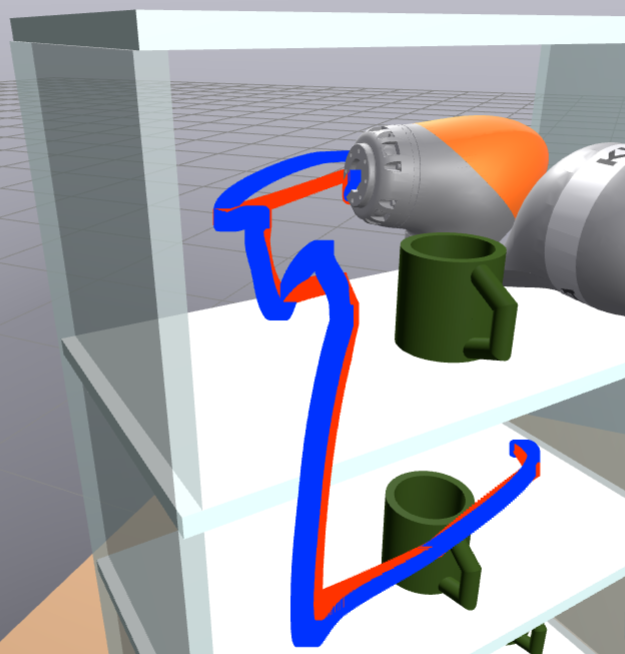}
    \end{subfigure}
    \hspace{3pt}
    \begin{subfigure}[b]{0.48\linewidth}
        \centering
        \includegraphics[width=\linewidth]{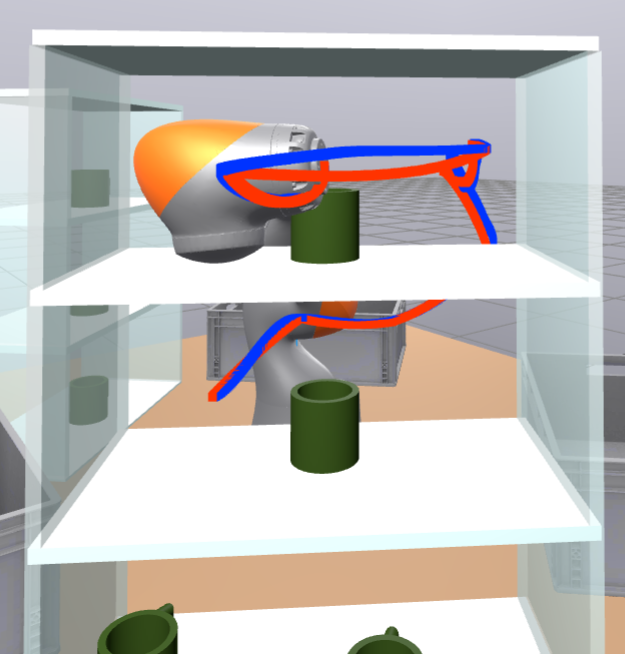}
    \end{subfigure}
    \caption{\change{A 7DoF iiwa reaching among shelves. Re-optimizing improves the path in the large region (in the top shelf), but shows minimal change for segments through smaller regions.} \tccomment{This seems cropped pretty aggresively. Any chance you can show more on either side?}}
    \label{fig:clutter}
\end{figure}

An obvious limitation of the proposed method is added computation time. We use a Python based custom PGD; \change{Commercial solvers, compiled languages, and performance optimization will speed up a mature implementation. This post-processing step will certainly be worth the additional runtime in cases like surgical robots that require strong guarantees and high quality. We have focused our numerical results on sparse environments. Our method works in dense clutter (see \Cref{fig:clutter}) given IRIS regions generated using new methods that better scale with environment complexity \cite{iriszo}. But the regions are smaller, leaving less room for improvement. In many contexts, robots move through areas of dense and sparse clutter, and our method can improve segments in the sparser regions, without adding collisions or worsening the trajectory in the densely-cluttered areas.}

Future work \change{could} include larger scale parallelization, especially if we integrate our post-processing step into the rounding stage. cuRobo \cite{curobo_icra23} has shown incredible results by solving many nonconvex trajectory optimization problems in parallel. This step could also be used in an Anytime Motion Planning framework \cite{mishani2023constant} where the later parts of a trajectory are refined as the earlier parts are traversed. Another possibility is using the nonconvex objectives with incremental search methods such as GCS*~\cite{gcsstar} \change{and Multi Query Shortest Path Problem in GCS~\cite{mqspp}.} Lastly, we \change{work on} designing better convex surrogates which still play an important role during the convex relaxation and initialization stages. Under clearly deficient convex surrogates (such as in the original constrained bimanual case), one can try to hand-design a better surrogate or potentially generate them automatically using learning-based approaches.

\bibliographystyle{IEEEtran}
\bibliography{IEEEabrv,ref}

\end{document}